\documentclass{svproc}
\usepackage[utf8]{inputenc}

\usepackage{graphicx}
\usepackage{amsfonts}
\usepackage{amsmath}
\usepackage{algorithm2e}
\begin{document}

\title{Towards Automatic Bayesian Optimization: A first step involving acquisition functions}

\author{Luis C. Jariego P\'erez \and Eduardo C. Garrido Merch\'an
}

\institute{Universidad Aut\'onoma de Madrid, Madrid, Spain\\
\email{luis.jariego@estudiante.uam.es}\\
\email{eduardo.garrido@uam.es}}
\maketitle

\begin{abstract}
Bayesian Optimization (BO) is the state of the art technique for the optimization of black boxes, i.e., functions where we do not have access to their analytical expression nor its gradients, are expensive to evaluate and its evaluation is noisy. A BO application is automatic hyperparameter tuning of machine learning algorithms. BO methodologies have hyperparameters that need to be configured such as the surrogate model or the acquisition function (AF). Bad decisions over the configuration of these hyperparameters implies obtaining bad results. Typically, these hyperparameters are tuned by making assumptions of the objective function that we want to evaluate but there are scenarios where we do not have any prior information. In this paper, we propose an attempt of automatic BO by exploring several heuristics that automatically tune the BO AF. We illustrate the effectiveness of these heurisitcs in a set of benchmark problems and a hyperparameter tuning problem. 
\end{abstract}

\section{Introduction}
Optimization problems, which task assuming minimization is to retrieve the minimizer $\mathbf{x}^* = \arg \max f(\mathbf{x}) \quad | \quad f : \mathbb{R}^n \to \mathbb{R}, \quad \mathbf{x}^*, \mathbf{x} \in \mathcal{X} \in \mathbb{R}^n$, are often solved easily when we have access to the gradient of the function that we want to optimize. Nevertheless, there exist a plethora of scenarios where we do not have access to these gradients. Typically, metaheuristics \cite{glover2006handbook} like genetic algorithms \cite{davis1991handbook} are used in this setting. Genetic algorithms and metaheuristics in general are useful when the evaluation of the function is cheap whether the cheap definiton refers to computational time or other resources such as the budget of the optimization process. This is not always the case. For example, we may consider an scenario when the function to optimize requires to configure a robot \cite{calandra2014bayesian} or training a deep neural network \cite{garrido2019predictive}. We can not afford in these scenarios a high number of evaluations. Ideally, we would like to consider a method that suggest as an approximation $\hat{\mathbf{x}}^* \approx \mathbf{x}^*$ of the optimum of the problem in the least number of evaluations as possible. An approximated solution to the true minimizer of the problem would be one with low absolute regret at the end of the optimization process $r = |f(\hat{\mathbf{x}})-f(\mathbf{x})|$, i.e. a local optima, not necessarily close, w.r.t. some distance metric in $\mathbb{R}^n$, in the input space to the minimizer. 

Moreover, we can even consider a more complicated scenario that the one described if the function that we want to optimize $f(\cdot)$ is modelled as a latent variable that we cannot observed as it has been contaminated by some random variable, for example, a gaussian random variable, hence observing $y = f(\cdot) + \mathcal{N}(0, \sigma)$ where $\sigma$ is i.i.d. $\forall \mathbf{x} \in \mathcal{X}$. In other words, for any two similar points of the input space we observe a, without loss of generality, gaussian distribution $\mathcal{N}(0, \sigma)$. Functions whose analytical expression is unknown, the evaluations are costly and the observation is contaminated with noise are often referred to as black boxes. Non convex Black box optimization has been dealt with success by BO methodologies \cite{brochu2010tutorial}, being the current state of the art approach. 

The most popular example of such an optimization is the task of automatic Machine Learning tuning of the hyperparameters or the hyperparameter problem of machine learning algorithms \cite{snoek2012practical},  such as the PC algorithm \cite{cordoba2018bayesian}, but also all kinds of subjective tasks like Suggesting Cooking Recipes \cite{garrido2018suggesting} or other applications belonging to robotics, renewable energies and more \cite{shahriari2015taking}.

Automatic Hyperparameter Tuning of Machine Learning algorithms is a desirable process that BO can tackle, but the BO procedure also have hyperparameters that need to be fixed a priori. As we are going to see in more detail in the next section, BO needs to fit a probabilistic surrogate model $M$, such as a Gaussian Process (GP) \cite{rasmussen2003gaussian}, in every iteration to the observations. This GP or other model have a set of hyperparameters $\theta$ associated with it. An Acquisition Function (AF) $\alpha(\cdot) : \mathbb{R}^n \to \mathbb{R}^n$ is then built in every iteration from the GP, or other model, that tries to represent an optimal tradeoff between the uncertainty given by the probabilistic model in every point of the input space and its prediction. The AF is a free hyperparameter of BO and it could be a bad choice depending on the problem. There are an infinite number of AFs $\alpha \in \mathcal{A}$, being $\mathcal{A}$ the functional space of possible AFs. There is no single AF that is the best for every problem. A bad choice on these and other hyperparameters of BO lead to bad results in the optimization process. Hence, we ideally need a process that performs automatic BO without the need of also hyperparametrize the BO algorithm. This work tries to attempt this problem and starts dealing with the automatic decision of which AF should we use by performing different heuristics. We hypothesize that an automatic BO algorithm will deliver better results than having to manually tune the hyperparameters of BO in problems where we do not have prior information about them. 

This paper is organized as follows, in section 2 we introduce the fundamental theory of BO and GP. Then, in section 3, we exhibit our proposed approaches for BO. We introduce a set of benchmark experiments and a real experiment to show the utility of our approach in an experiments section. Finally, a conclusions and further work section summarizes the paper.

\section{BO Issues for Automatic Optimization}
The BO algorithm is executed in an iterative fashion, where it uses a probabilistic surrogate model $M(\theta)$ as a prior over functions $p(F)$ which functional space $\mathcal{F}$ contains all the hypotheses about the objective function $f(\cdot)$ that we want to get the maximum of $\mathbf{x}^* = \arg \max f(\mathbf{x})$. This model $M$, hyperparametrized by a set $\theta$, is typically a GP \cite{rasmussen2003gaussian}, but other models such as Bayesian Neural Networks \cite{springenberg2016bayesian} and Random Forests \cite{kotthoff2017auto} are also used. In order for BO to work, we need to assume that the function $f$ can be sampled from it $p(F)$. Hence, depending on the problem, different models may be optimal and even some of them may led to bad result, being hence the model and its hyperparameters a hyperparameter of BO. For example, if we consider the popular GP for a problem, if the objective function is not stationary and we do not do any transformation of the input space to treat this property of the objective function, the GP does not serve as a prior for that function and independently of the other hyperparameters of the BO algorithm and of the number of evaluations, we are going to retrieve bad results. 

Even by choosing the same probabilistic surrogate model $M$ we need to define the correct hyperparameters $\theta \in \Theta$ for that model. In the typical case of a GP, a wrong choice of kernel can imply that the function that we want to optimize is no longer on the functional space that the GP defines. Even by optimizing the rest of the GP hyperparameters by a maximum likelihood procedure or taking an ensemble of different GPs with hyperparameters sampled from a hyperparmeter distribution, as they depend on the choice of kernel, that optimization procedure would be useless, leading again the BO algorithm to bad results.

BO uses the predictive distribution of the  model in every point $\mathbf{x}$ of the input space $\mathcal{X}$ to build an AF $\alpha(M(\mathcal{X}|\theta))$. This AF represents the utility of evaluating every point $\mathbf{x} \in \mathcal{X}$ in order to retrieve the optimum of the objective function in the, in the standard BO algorithm, next step of the iteration, being a myopic optimization procedure. The literature contains different AFs that try to represent the optimal trade off between exploration of the space areas that have not been yet explored and the exploitation of previously good evaluated results. Some of these AFs are the following ones:

Probability of Improvement: 
$\mathrm{PI}(\boldsymbol{\mathrm{x}}) = \Phi(\dfrac{f(\boldsymbol{\mathrm{x}}_{best}) - \mu(\boldsymbol{\mathrm{x}})}{\sigma(\boldsymbol{\mathrm{x}})}).$ This AF basically represents, for each point of the space, the probability of this point to be better if evaluated than the best observed value retrieved so far.

Expected Improvement: 
$\mathrm{EI}(\boldsymbol{\mathrm{x}}) = \sigma(\boldsymbol{\mathrm{x}}) (\gamma(\boldsymbol{\mathrm{x}}) \Phi(\gamma(\boldsymbol{\mathrm{x}})) + \phi(\gamma(\boldsymbol{\mathrm{x}}))).$. The previous function does not take into account, for every point and sample function of the probabilistic model, how much does the point improve the maximum value found. Expected improvement represents a theoretical improvement over the probability of improvement by considering this quantity.

Lower Confidence Bound: 
$\mathrm{LCB}(\boldsymbol{\mathrm{x}}) = \mu(\boldsymbol{\mathrm{x}}) - \kappa \sigma(\boldsymbol{\mathrm{x}}).$. This AF is representing a tradeoff between the prediction of the probabilistic model in each point of the space $\mu(\boldsymbol{\mathrm{x}})$ and exploration over unknown areas given by the uncertainty of the model in each point of the space $\sigma(\boldsymbol{\mathrm{x}}$. The $\kappa$ parameter assigns a weight for each quantity.

But there are a lot more, in fact, we could generate an infinite number of possible AFs. As in the case of the probablistic surrogate model, the decision of the chosen AF conditions the optimization. For example, if the function is monotonic, we do not need a heavy exploratory based AF. On the other way, if the objective function is contaminated by a high level of noise, the exploitation criterion is practically useless. There is no single best AF for every scenario, as the no free lunch theorem states \cite{ho2002simple}.

\begin{algorithm}[H]
\label{alg:bo}
 \caption{BO of a black-box objective function.}
 \For{$\text{t}=1,2,3,\ldots,\text{max\_steps}$}{
        {\bf 1:} Find the next point to evaluate by optimizing the AF:
        $\mathbf{x}_t = \underset{\mathbf{x}}{\text{arg max}} \quad \alpha(\mathbf{x}|\mathcal{D}_{1:t-1})$.

        {\bf 2:} Evaluate the black-box objective $f(\cdot)$ at $\mathbf{x}_t$: $y_t=f(\mathbf{x}_\text{t}) + \epsilon_t$.

        {\bf 3:} Augment the observed data $\mathcal{D}_{1:t}=\mathcal{D}_{1:t-1} \bigcup \{\mathbf{x}_t, y_t\}$.

        {\bf 4:} Update the GP model using $\mathcal{D}_{1:t}$.
 }
 \KwResult{ Optimize the mean of the GP to find the solution. }
 \vspace{.25cm}
\end{algorithm}

BO have more hyperparameters, as for example the optimization algorithm of the AF, typically a grid search over the space of the AF and a local optimization procedure such as the L-BFGS algorithm \cite{gao2004improved}. The sampling procedure for the hyperparameter distribution of the probabilistic surrogate model, the number of samples and more. Varying the values of those hyperparameters condition the quality of the final recommendation. We have observed that despite the fact that BO is an excellent optimization procedure, it is not automatic and we need to choose wisefully the hyperparameters. This is possible if we have prior knowledge about the objective function, but this is not a scenario that always happens.

Hence, we ideally need a procedure to search for the best BO hyperparameters, concretely the model and the acquisition, as the function is being optimized. This work is a first step towards this goal. We explore different simple heuristics to determine if they affect to the optimization behaviour. We have only focused on the AFs, but the selection of a particular probabilistic surrogate model while the optimization is being performed is also an essential issue to deliver automatic BO.  

The next section will illustrate the first possible methods that we can execute to perform a simple search of the possible AFs belonging to the set $\mathcal{A}$ of all possible AFs to build from a probabilistic surrogate model.

\section{Heuristic driven Bayesian Optimization}

In this work, we begin to explore the possibilities of combining AFs in order to build criteria that satisfies the majority of the problems or that it adapts to the optimization process.

Formally, if we have a set $\mathcal{A}$ of AFs, we are going to build criteria that combines these AFs.

We hypothesize that different GP states of an underlying objective function need different AFs in order to discover which is the optimum of the underlying function. Which is in contrast to the typical BO algorithm that just uses the same AF for all the iterations.

We propose, given the same probabilistic surrogate model, using different AFs or linear combinations between AFs in the same BO algorithm. For every iteration, a different AF will be used, defining now for BO problems not an AF as in standard BO but an AF generator $\mathcal{G}$ that generates for every iteration $t=1..N$ a different AF $\alpha_t(\cdot) \in \mathcal{A}$. These generators can use any possible AF as seeds for the generation of AFs in every iteration. We illustrate different approaches for an AF generator that are basically heuristics that search the best possible AF.
 
In practice, we have explored combinations of Standard AFs used in the BO literature. We formulate the hyperparameter tuning of AFs for BO as a search problem and start tackling it with heuristics to observe how the global behaviour of BO is conditioned.
        
We propose the following approaches over the AFs described in the previous section. As it has been described, we could use an extended set of AFs like including PES \cite{hernandez2014predictive}, MES \cite{wang2017max} or any other. We also hypothesize that the behaviour of the heuristics will improve with the addition of more and more diverse AFs to the seed set of AFs that we consider. The heuristics that we propose are, in first place, the Random criteria, basically defined by placing un uniform distribution $\mathrm{U}$ over the functional set of AFs $\mathcal{A}$ and sampling from it in every iteration. For every iteration a different AF $\alpha(\cdot)_t$ is going to be executed. We hypothesize that the optimization process will be enriched by the random execution of different criteria, obtaining good results. In our case, as we only consider the EI, LCB and PI acquisitions, the criterion will be given by the following expression: 
$Rand(\boldsymbol{\mathrm{x}}) = \mathrm{U} (\mathrm{PI}(\boldsymbol{\mathrm{x}}), \mathrm{EI}(\boldsymbol{\mathrm{x}}), \mathrm{LCB}(\boldsymbol{\mathrm{x}})).$, but in the general case it would be: $Rand(\boldsymbol{\mathrm{x}}) = \mathrm{U} (\mathcal{A}).$

We could perform the same logic as in the Random case but performing a Sequential criterion. $Seq(\boldsymbol{\mathrm{x}}, n_{iter}) = Cands(\boldsymbol{\mathrm{x}})[ n_{iter}  mod( n_{cands} ) ].$. We model here all the acquisitions in an ordered list and sample them sequentially, one acquisition for every iteration. We have proposed this two initial strategies in an analogy with respect to the grid search and random search, hypothesizing that they fully explore the set of seed AFs and enriching the optimizing process results.

If we assume that all the AFs can be valid in any time of the optimization process and retrieve different but interested results, then, a logical suggestion will be to consider a linear combination over all the considered AFs, that is the weighted AF criterion, defined by the following expression: $\alpha_w(\mathbf{x}|\mathcal{A},\mathbf{w}) = \sum_{i=1}^{|\mathcal{A}|} w_i\alpha_i(\mathbf{x}) : \sum_{i=1}^{|\mathcal{A}|} w_i = 1$. In our  particular case the weighted criterion function would be $\alpha_w(\boldsymbol{\mathrm{x}}) = \kappa_{\mathrm{PI}}  \mathrm{PI}(\boldsymbol{\mathrm{x}}) + \kappa_{\mathrm{EI}}  \mathrm{EI}(\boldsymbol{\mathrm{x}}) + \kappa_{\mathrm{LCB}}  \mathrm{LCB}(\boldsymbol{\mathrm{x}}).$

Lastly, lots of metaheuristics and machine learning algorithms include mechanisms such as the mutation probability in genetic algorithms or dropout in deep neural networks that act as regularizers, enforcing exploration and preventing from overfitting, improving the results. We hypothesize that we can establish an analogy for the AF search so we introduce a noised criterion, that basically transforms the acquisition in a latent functional variable and contaminates it with i.i.d gaussian noise to enforce exploration: $f(\boldsymbol{\mathrm{x}}) = g(\boldsymbol{\mathrm{x}}) + acquisition\_noise  \mathcal{N}(0,\mathbf{I}).$
        
All these approaches are heuristic but explore a space defined by the set $\mathcal{A}$.
Our procedure combines AFs like this: The weighted AF criterion contains a weight for each AF to measure its the importance. This is a generalization of common BO but does not solve the automatic BO scenario. If, instead of being hardcoded by the user, these weights were adapted as the problem is being optimized or in function of the problem, the optimization would be automatic. As a first attempt towards automatic BO, we propose to use a Metaoptimization of the weights $\mathbf{w}$ using BO over the weight space $\mathcal{R}^{|\mathcal{A}|} \in [0,1]^{|\mathcal{A}|}$. We define a search space of $|\mathcal{A}|$ weights that are associated with their respective AFs. Then, we execute a standard BO procedure that gives us the weights that minimize the predicted error by the underlying BO algorithm. By performing this double loop, the weights are optimized and the underlying BO algorithm is automatic. Nevertheless, the upper BO algorithm still needs to be tuned but we can study several problems to adjust a reasonable prior over the weight space.
        
\section{Experiments}
We carry out several experiments to evaluate the performance of the described heuristics in the previous section. We also compare the approaches to a pure exploration method based on Random Search \cite{bergstra2012random}. The set of seeds AFs and the proposed ones have been implemented in SkOpt \cite{markov2017skopt}. In each experiment carried out in this section we report average results and the corresponding standard deviations. The results reported are averages over 100 repetitions of the corresponding experiment. Means and standard deviations are estimated using 200 bootstrap samples. The hyperparameters of the underlying GPs are maximized through maximum likelihood in the optimization process. The AF of each method is maximized through a grid search. 

\subsection{Benchmark Experiments}
\begin{figure}[htb]
\begin{center}
\includegraphics[width=0.7\linewidth]{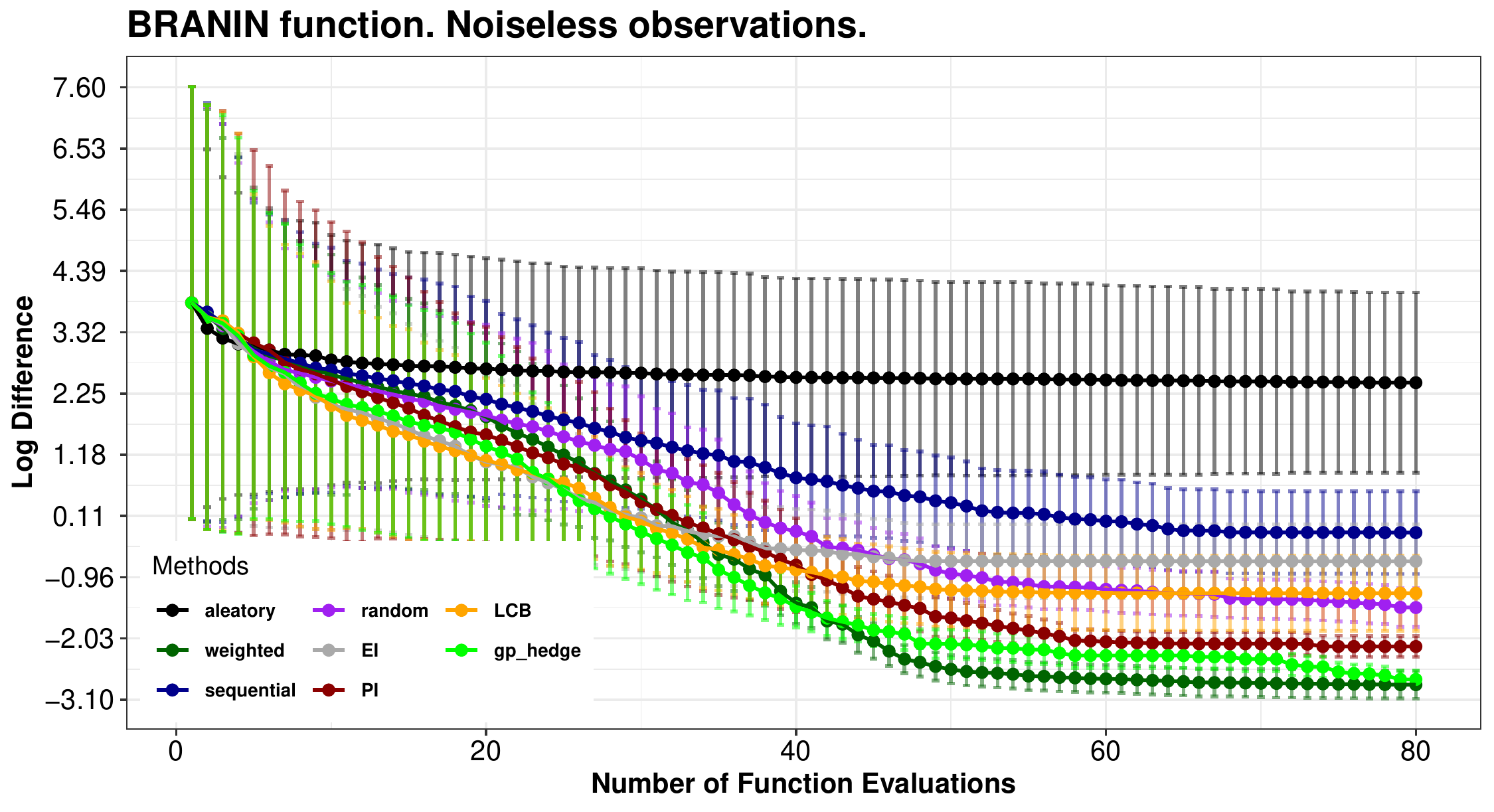}
\end{center}
\caption{Means and standard deviations of the log difference w.r.t the absolute regret of the maximizer of the different considered AFs in the Branin Function.}
\label{fig:bench}
\end{figure}

\begin{figure}[htb]
\begin{center}
\includegraphics[width=0.7\linewidth]{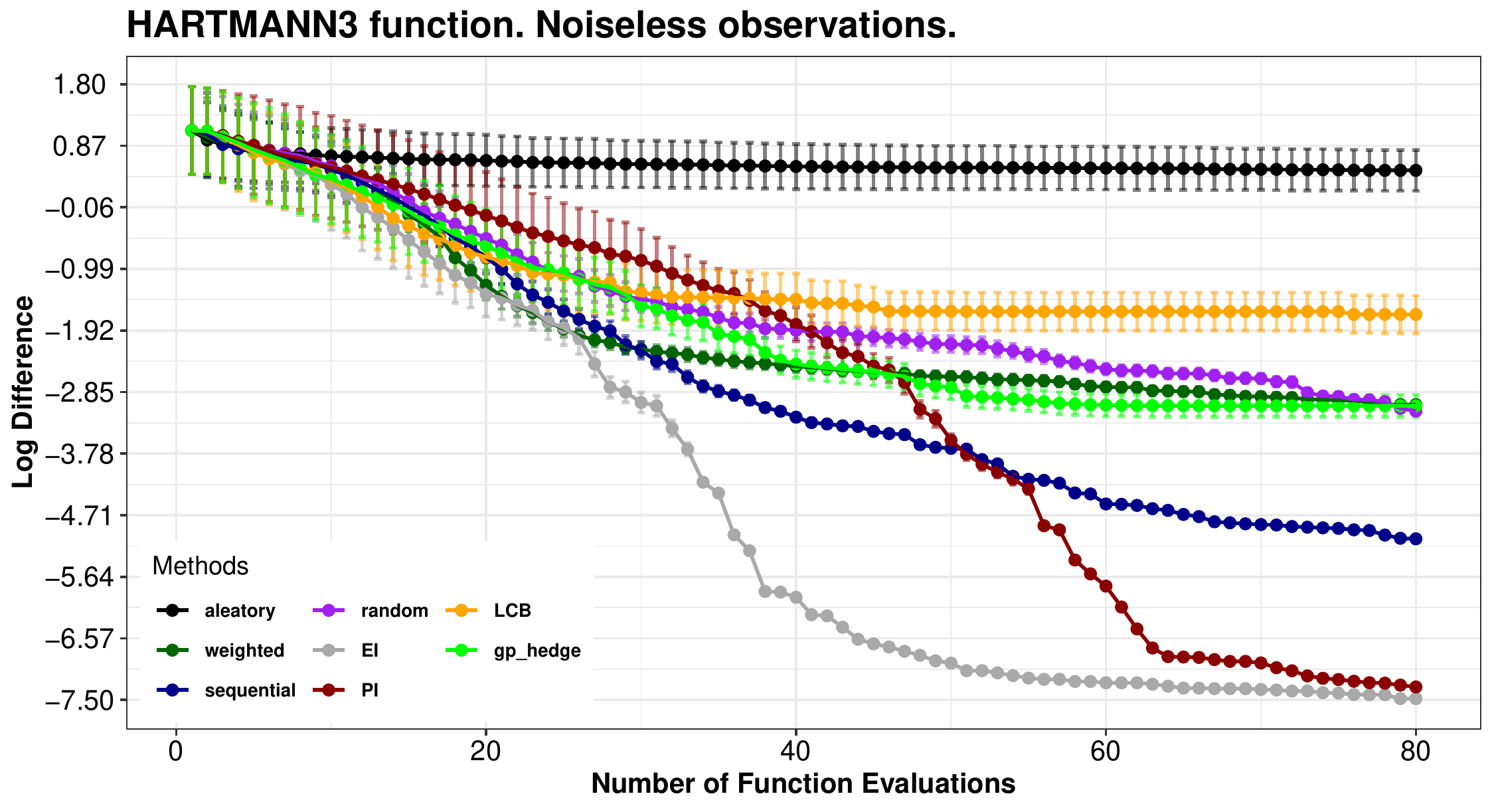}
\end{center}
\caption{Means and standard deviations of the log difference w.r.t the absolute regret of the maximizer of the different considered AFs in the Hartmann Function.}
\label{fig:bench_2}
\end{figure}

\begin{figure}[htb]
\begin{center}
\includegraphics[width=0.7\linewidth]{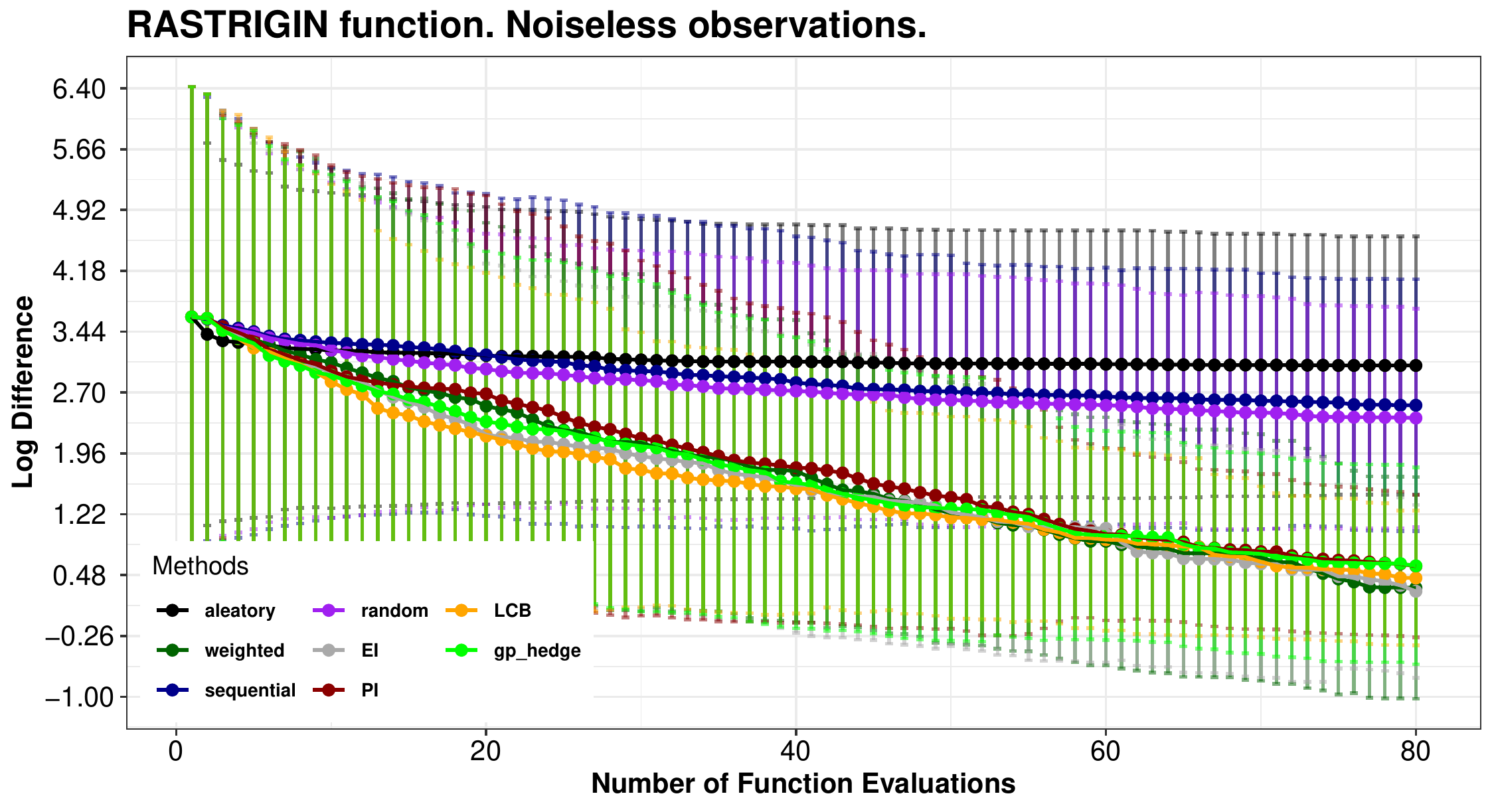}
\end{center}
\caption{Means and standard deviations of the log difference w.r.t the absolute regret of the maximizer of the different considered AFs in the Rastrigin Function.}
\label{fig:bench_3}
\end{figure}

We test the proposed AFs and compare with GP-Hedge over a set of benchmark problems, namely, the Branin, 3-dimensional Hartmann and 3-dimensional Rastrigin functions. We plot the results in Figures \ref{fig:bench}, \ref{fig:bench_2} and \ref{fig:bench_3}.

We can observe that, for the Branin function, the best method is the weighted AF optimized by the metaoptimization process. GP-Hedge method also delivers good results, tying at the end with the weighted AF. We hypothesize that the good behaviour of the ensemble AFs (weighted and hedge) is a consequence given by the fact that every seed adds some value in the problem. Separated, although, they do not provide good results.

We observe a different behaviour in the Hartmann function, where only the pure exploitation AFs (EI and PI) report a good result. This happens due to the shape of Hartmann, where exploration is a bad strategy as with pure exploitation we can reach to the optimum. We can observe empirically that EI is better than PI as it considers the amount of improvement over the incumbent. Ensemble AFs, as they consider exploration or other criteria rather than EI and PI lose performance, but they are not as bad as LCB, which is not a good strategy here. This property of ensemble AFs guarantees that they are not as bad as the worst case in any scenario.

In the Rastrigin function, we can observe that the random methods do not perform well but the others tie, performing a better result. No AF seems to govern, maybe all locating just local optima of Rastrigin. The large standard deviations of the Rastrigin function may be explained for different reasons, first is the shape of the function with lots of local optima, each repetition may end in different points and hence the deviation is big. Other explanations are  the optimization of the AF being done with a grid search. We need to perform a L-BGFS optimization of the maximum valued point retrieved by this search to discard the hypothesis that the large deviations are happening for local optima. Another important fact is to consider a hyperparameter distribution of the GPs to sample from it with an algorithm such as slice sampling instead of simply optimizing the hyperparameters through maximum likelihood, incurring in overfitting of the model as BO performs a small number of evaluations.

\subsection{Real Experiment}                
\begin{center}
\begin{figure}[htb]
\begin{center}
\includegraphics[width=0.57\linewidth]{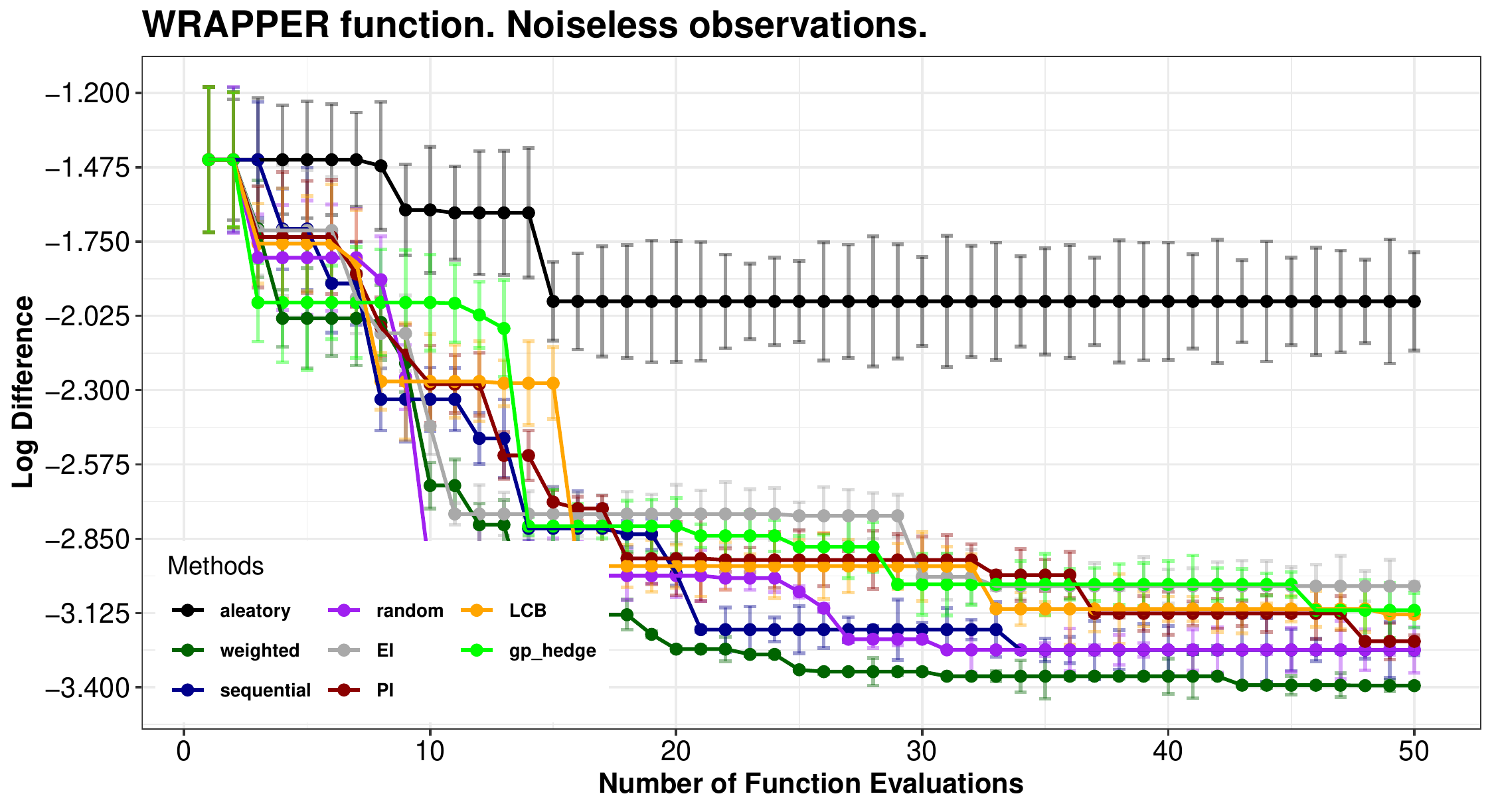}
\end{center}
\caption{Means and standard deviations of the log difference w.r.t a perfect classification error of the different considered AFs in the Hyperparameter Tuning of a Gradient Boosting Ensemble.}
\label{fig:real}
\end{figure}  
\end{center}

In this section we perform a hyperparameter tuning problem of the learning rate, minimum samples split and maximum tree depth of a Gradient Boosting Ensemble classifier on the Digits Dataset. We do not find the issues of the Rastrigin function in this problem as, typically, the shape of the estimation of the generalization error function for machine learning algorithms is smooth, so we expect that the retrieved results by BO in this case will not contain a high standard deviation and favour the weighted criterion. The results can be seen in Figure \ref{fig:real}.

As we can see, the weighted criterion is the best one in this problem, that might contain some local optima and irregularities as the random search also work pretty well. Maybe due to certain combinations of parameters that generates good results. There is a lot more to do for automatic BO but the first necessary step towards that goal is to explore the set of all possible AF through, as in this case, generators of linear combinations of AFs that, in average, produce great results.

\section{Conclusions and Further Work}
The proposed approaches provide alternatives for Hyperparameter Tuning problems with respect to the standard AFs. There is still a lot of work to do for automatic BO, such as doing a similar approach as this one but with probabilistic graphical models and AF optimizers.
In future work, we would like to build a dataset from a plethora of GP states and try to train a deep neural network that learns to predict which is the best AF to use or even the best point to consider given the dataset and the state of the current GP. We would like to test whether if the
transformations made in the input space to deal with integer \cite{garrido2017dealing} and categorical-valued variables
\cite{garrido2020dealing} change the behaviour of the given AF heuristics. The final purpose of this research is to employ automatic BO for the optimization of the hyperparameters of the machine learning architecture of the creative robots that exhibit human behaviour \cite{merchan2020machine} \cite{garrido2020artificial} to test machine consciousness hypotheses.
\section*{Acknowledgements}
The authors gratefully acknowledge the use of the facilities of Centro de Computaci\'on Cient\'ifica (CCC) at Universidad Aut\'onoma
de Madrid. The authors also acknowledge financial support from Spanish Plan Nacional I+D+i, grants TIN2016-76406-P and TEC2016-81900-REDT.
\bibliographystyle{acm}
\bibliography{main}
\end{document}